\documentclass{article}
\usepackage[accepted]{icml2021}
\icmltitlerunning{Reinforcement Learning for (Mixed) Integer Programming: Smart Feasibility Pump}

\usepackage{times}
\usepackage{soul}
\usepackage{url}
\usepackage[hidelinks]{hyperref}
\usepackage[utf8]{inputenc}
\usepackage[small]{caption}
\usepackage{subcaption}
\usepackage{graphicx}
\usepackage{amsmath}
\usepackage{amsthm}
\usepackage{amssymb}
\usepackage{booktabs}
\usepackage{bm}
\usepackage{algorithm}
\usepackage{algorithmic}
\usepackage{xcolor}
\usepackage{microtype}
\DeclareMathOperator*{\argmin}{arg\,min}

\begin{document}

\twocolumn[
\icmltitle{Reinforcement Learning for (Mixed) Integer Programming: Smart Feasibility Pump}



\icmlsetsymbol{equal}{*}

\begin{icmlauthorlist}
\icmlauthor{Meng Qi}{equal,ucb}
\icmlauthor{Mengxin Wang}{equal,ucb}
\icmlauthor{Zuo-Jun (Max) Shen}{ucb}

\end{icmlauthorlist}

\icmlaffiliation{ucb}{Department of Industrial Engineering and Operations Research, UC Berkeley}

\icmlcorrespondingauthor{Meng Qi}{meng\_qi@berkeley.edu}
\icmlcorrespondingauthor{Mengxin Wang}{mengxin\_wang@berkeley.edu}

\icmlkeywords{Mixed-Integer Programming, Deep Reinforcement Learning}

\vskip 0.3in
]


\printAffiliationsAndNotice{\icmlEqualContribution} 


\begin{abstract}
Mixed integer programming (MIP) is a general optimization technique with various real-world applications.
Finding feasible solutions for MIP problems is critical because many successful heuristics rely on a known initial feasible solution. However, it is in general NP-hard. 
In this work, we propose a deep reinforcement learning (DRL) model that efficiently finds a feasible solution for a general type of MIPs. In particular, we develop a smart feasibility pump (SFP) method empowered by DRL, inspired by Feasibility Pump (FP), a popular heuristic for searching feasible MIP solutions. 
Numerical experiments on various problem instances show that SFP significantly outperforms the classic FP in terms of the number of steps required to reach the first feasible solution. We consider two different structures for the policy network. The classic perception (MLP) and a novel convolution neural network (CNN) structure. The CNN captures the hidden information of the constraint matrix of the MIP problem and relieves the burden of calculating the projection of the current solution as the input at each time step. This highlights the representational power of the CNN structure.
\end{abstract}

\section{Introduction}

Integer programming (IP) and mixed integer programming (MIP) are mathematical optimization problems where all or some of the decision variables are restricted to be integers. IPs and MIPs are critical for solving discrete and combinatorial optimization problems in various applications, including production planning, scheduling, and vehicle routing (we refer to \cite{pochet2006production}, \cite{sawik2011scheduling}, and \cite{malandraki1992time} for more details). 
The study of theory and computation methods dates back to several decades ago (\cite{dantzig1954solution}, \cite{land2010automatic}, and \cite{padberg1973facial}).

Despite the importance of solving IPs and MIPs in application, they are generally very difficult to solve in theory (NP-hard) and in practice. There is no polynomial time algorithm that can guarantee to solve a general IP/MIP. Besides exact methods (Branch-and-bound, branch-and-cut, etc.), heuristic algorithms are widely adopted for their simplicity and efficiency. Many successful heuristics including local branching \citep{fischetti2003local} and RINS \citep{rins}, require an initial feasible solution to start with. Finding a feasible solution is often the first step and one of the most critical steps of solving IP/MIPs. However, it is also NP-hard in general; even well-developed commercial solvers, such as CPLEX, may struggle or even fail.

In this study, our goal is to provide a deep reinforcement learning (DRL) model that efficiently finds a feasible solution for a general type of IP/MIPs. In particular, we consider IP/MIP problems with a linear objective, linear constraints, and integral constraints. Such formulation is quite general since any combinatorial optimization problem with finite feasible regions can be reformulated as an IP/MIP in this form \citep{tang2020reinforcement}. We aim at developing a DRL agent that smartly perform the idea of \textit{Feasibility Pump}, which is one of the most well-known algorithms for finding a feasible solution for MIPs in optimization theory.

The main idea of FP is to decompose the original problem into two problems: one focuses on the feasibility of a continuous relaxation and the other retains integrality. The original work of FP was introduced by \cite{fischetti2005feasibility} for $0-1$ MIPs. The authors demonstrated that FP is very effective in finding feasible solutions.
Later, \cite{bertacco2007feasibility} extended this algorithm to general MIPs and introduced a random component in the rounding function.
Further extensions of FP include the following. 
\cite{fischetti2009feasibility} adjusted the rounding method so that a feasible solution can be found in fewer steps. \cite{baena2011using} and \cite{boland2014boosting} investigated the idea of using integral reference points to further improve the rounding steps. \cite{achterberg2007improving} proposed the objective feasibility pump, which includes the original objective in the objective function of FP to improve the quality of the feasible solution. \cite{boland2012new} proposed a penalty system for non-integrality to prevent the FP from cycling. \cite{bonami2009feasibility} and \cite{bonami2012heuristics} extended the FP algorithm to nonlinear MIPs. 

 Although this heuristic is widely adopted, it requires solving an optimization problem within each iteration, which is especially inefficient when extending to nonlinear cases. The efficiency may also be weakened because FP may loop between infeasible points.

Recently, several studies have been conducted exploring the topic of using machine learning methods to solve IP/MIPs. 
The main drawback of adopting supervised learning schemes is that they require known (near) optimal solutions in the training set, which is difficult to achieve for IP/MIPs. 
One of the most closely related work is \cite{fischetti2017deep}, where they trained deep neural networks to find a feasible solution for 0-1 MIPs.

Some other works have aimed to use machine learning methods for branch-and-bound algorithms.
\cite{he2014learning} used imitation learning to learn an adaptive node searching order to achieve better solutions faster for MIPs.
\cite{learning_branch} learned an easy-to-evaluate surrogate function that mimics the strong branching strategy for branch-and-bound. \cite{nair2020solving} proposed the methods of neural diving and neural branching, to improve the high-quality joint variable assignment and bounding the gap, respectively.

Other related works have studied combinatorial problems, which are a subset of IPs. The most related works include \cite{khalil2017learning}, where the authors proposed a method based on RL to solve various combinatorial problems on graphs. \cite{bello2016neural} proposed a framework based on recurrent neural network (RNN) and deep reinforcement learning (DRL) to general combinatorial problems. 
\cite{learn_vrp} proposed a simplified version of RNN and DRL to tackle the vehicle routing problem (VRP). \cite{li2018combinatorial} combined deep learning techniques with a graph convolutional network to solve combinatorial problems based on graphs. \cite{delarue2020reinforcement} investigated VRP and developed a framework for a value-function-based RL model where the action selection problem is formulated as a mixed-integer optimization problem. \cite{tang2020reinforcement} designed RL methods to enhance the cutting plane method for solving IPs.

So far to our knowledge, there is no existing work that focuses on feasibility study for general MIPs using RL models. We believe that RL models can be utilized to learn feasible solutions for a general type of MIPs.

\textbf{Our contributions.} 
In this work, we propose a general framework, the \textit{smart feasibility pump} (SFP) model, which utilizes the RL frameworks based on the spirit of FP. More specifically, we propose two different SFP models: SFP based on multi-layer perception (SFP-MLP) and SFP based on convolutional neural network (SFP-CNN), 
both find feasible solutions for IP/MIPs more efficiently, that is, with fewer search steps. Moreover, to better capture the structure of the constraint matrix, SFP-CNN adopts a novel CNN structure for policy learning networks that significantly improves the performance of SFP-CNN compared to SFP-MLP and the original FP.
We summarize the contribution of our work as follows:
\begin{itemize}
    \item \textbf{A RL model for feasible solutions of MIPs.} 
    Different from existing literature which mostly focus on one specific application and aim for the optimal solution, our work is the first attempt to use (deep) RL methods for seeking feasible solutions for a class of general MIPs. 
    
    \item \textbf{The spirit of a successful heuristic.} Our model is inspired by FP, one of the most popular heuristic algorithms for finding feasible solutions for IP/MIP. The effectiveness of decomposing the continuous relaxation and integrality has been verified by the success of FP.
    Adopting a similar idea, we propose the smart feasibility pump model because it is empowered by deep RL models.
    
    \item \textbf{A novel CNN for constraint matrix.} Besides a regular MLP, we innovatively adopt a convolutional structure for the policy network to capture the structure of constraint matrix of MIPs. The CNN provides us with a stronger representational power to capture the hidden structure of the constraint matrix.
    
    \item \textbf{Empirical evaluation.} We conduct numerical experiments with different problem dimensions and number of constraints. The results demonstrate the significant advantages of the SFP models compared to the original FP.
\end{itemize}

\section{Background}
\label{sec: background}
\textbf{Mixed Integer Programming.}
In this work, we aim at finding a feasible solution of a generic Mixed Integer Programming (MIP) problem. A generic MIP problem can be written as follows:
\begin{subequations}
\label{mip}
\begin{align}
\min & \quad c^Tx \label{mip:obj}\\
s.t. &\quad  Ax \leq b  \label{mip:cons}\\
 & x_i \in \mathbb{Z}, \forall i\in S \label{mip:int}
\end{align}
\end{subequations}

where $A \in \mathbb{R}^{m\times n}$, $c,b \in \mathbb{R}^n$, and $x \in \mathbb{R}^n$ denote the decision variables. A subset of the decision variables is integral, and $I$ denotes the index set of the integer variables. When $I = \{1,2,...,n\}$, it becomes an IP problem. The MIP problem aims to minimize a linear objective function \eqref{mip:int} while satisfying a set of linear constraints \eqref{mip:cons} and the integral constraints \eqref{mip:int}. A feasible solution is an element in the feasible region $\mathcal{P} = \{x\in \mathbb{R}|Ax\leq b, x_i \in \mathbb{Z}, \forall i \in S\}$. To avoid triviality, we focus on the case when $\mathcal{P}$ is bounded and excludes the origin.

Solving an MIP problem is generally known to be NP-hard. Even well-developed commercial solvers (for instance, Cplex, Gurobi, and Mosek) may struggle or fail. Many successful heuristics for solving MIPs, such as local branching \citep{fischetti2003local} and RINS \citep{rins}, only work with a known initial feasible solution. It is of great importance to find a good feasible solution for MIPs. However, finding a feasible solution for a MIP is also NP-hard, owing to the well-known equivalence of the optimization problem and the feasibility problem in terms of computational complexity. 

\begin{figure}[h]
\centering
  \includegraphics[width=0.55\linewidth]{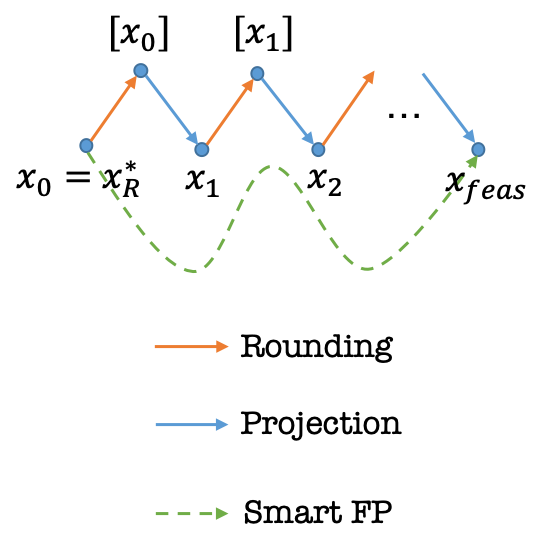}
  \caption{Illustration of the Feasibility Pump}
  \label{fig:fp}
\end{figure}

\textbf{Feasibility Pump.} The feasibility pump is one of the most popular approaches for finding the initial feasible solution of MIPs. The basic idea of the FP algorithm is to iteratively find and round a continuous relaxation solution for the MIP. Let $\mathcal{P}_R = \{x \in \mathbb{R}|Ax\leq b\}$ denote the continuous relaxation of $\mathcal{P}$. The overall procedure is shown in Algorithm \ref{alg:fp} and illustrated in Figure \ref{fig:fp}. The FP algorithm starts with the rounded optimal continuous relaxation solution of the MIP and then searches for the nearest points in the relaxed feasible region. It continues perturbing and rounding the new point found at each step until a feasible solution is discovered or the limit of maximum number of steps is reached. Despite being a powerful heuristic, it requires solving an optimization problem within each iteration, which becomes especially inefficient when the problem size increases or extends to nonlinear constraint cases. In addition, the FP algorithm tends to converge to and loop between infeasible points with non-binary integer variables \citep{fischetti2005feasibility}. 

\begin{algorithm}[H]
\begin{algorithmic}[1]
 \STATE Initialization: $x^0 \leftarrow \argmin_{x\in \mathcal{P}_R}c^T x$; $\bar{x}^0\leftarrow [x^0]$; $k\leftarrow 0$\\
 \WHILE{$\bar{x}^k$ is not feasible}
 \STATE $x^{k+1}\leftarrow\argmin_{x\in \mathcal{P}_R}\|x - \bar{x}^k\|$\;
 \STATE $\bar{x}^{k+1} \leftarrow [x^{k+1}]$
 \IF{$\bar{x}^{k+1} = \bar{x}^k$}
  \STATE random perturbation of $\bar{x}^k_j, \forall j\in I$.
  \ELSE 
  \STATE $k \leftarrow k + 1$
  \ENDIF
 \ENDWHILE
 \STATE Return $\bar{x}^k$
\end{algorithmic}
 \caption{Feasibility Pump. With a slight abuse of notation, $[x]$ denotes the rounding of $x$, i.e. $[x]_i = [x_i],\forall i \in I$ and $[x]_i = x_i \forall i \notin I$. The algorithm terminates if reaching the maximum number of iterations with no feasible solution found. }
 \label{alg:fp}
\end{algorithm}


\section{Smart Feasibility Pump: a RL Formulation}
Here, we present our formulation of the SFP method as an RL problem. We aim to learn an RL policy that can find a feasible solution for any randomly generated IP/MIPs with the form of (\ref{mip}). 
\subsection{RL Formulation}
In this section, we describe two slightly different formulations for the SFP-MLP and the SFP-CNN. 
For each of the formulations, we start by specifying the Markov decision process. At each time step $t$, we denote the state as $s_t \in \mathcal{S}$, the action as $a_t \in \mathcal{A}$, and the instant reward as $r_t\in \mathbb{R}$. A policy $\pi:\mathcal{S} \rightarrow \mathcal{P}(\mathcal{A})$ provides a mapping from any state to a distribution over actions $\pi(\cdot|s_t)$. Our goal is to learn a policy that maximizes the expected cumulative rewards over a horizon $T$, that is, $\max_\pi J(\pi) := \mathbb{E}_\pi [\sum_{t=0}^{T-1} \gamma^t r_t]$, where $\gamma \in (0,1]$ is a discount factor. In the remaining of this section, we specify the state space $\mathcal{S}$, action space $\mathcal{A}$, reward $r_t$, and the transition from $s_t$ to $s_{t+1}$.

\textbf{State Space $\mathcal{S}$ for SFP-MLP.} In the case of MIP, we let the state at each time $t$ to be $s_t = (\text{Flat}(A), b, x_t,\tilde{x}_t, I)$, $\tilde{x}$ denotes the projection of the current solution to the feasible region defined by constraints (\ref{mip:cons}). The idea of including a projection $\tilde{x}_t$ is borrowed from the original FP algorithm, which provides useful information for the agent about the direction to move to achieve feasibility. It improves the learning ability of the agent yet with the cost of large computational efforts at each time step. 
Moreover, $\text{Flat}(A)$ is a vector that equals the flattened constraint matrix $A$ in the original problem (\ref{mip:obj})-(\ref{mip:int}), and $I$ is a binary vector that indicates whether the $i$th variable is an integer (whether $i\in \mathcal{I}$). For IPs, as we know that all decision variables in (\ref{mip:obj})-(\ref{mip:int}) are integrals, we set the state vector as $s_t = (\text{Flat}(A), b, \tilde{x}, x)$. 

As explained later in Section \ref{sec: policy-learning}, if we use our proposed network structure with CNN, we can avoid calculating the projection of $x_t$ at each time step $t$ and instead use the initial solution $x_0$ as $\tilde{x}_t$ for all time steps.

\textbf{State Space $\mathcal{S}$ for SFP-CNN.} When we adopt our proposed CNN as the policy network, because of the ability of the CNN to capture the structure of the constraint matrix, the learning ability of the RL agent is brought up to a level that we can get rid of the projection of $x_t$ at each time step $t$. Instead, it only requires the projection to be performed once for the initial point $x_0$ and use it as $x_t$ for all $t>0$. Therefore, for MIPs, we defined the state vector as $s_t = ([A, b], x_t , \tilde{x}_0, I)$, where $[A, b]$ denotes the matrix defined as matrix $A$ concatenated with column vector $b$. Similarly, for IPs, we have $s_t = ([A, b], x_t , \tilde{x}_0)$.

\textbf{Action Space and State Transition.} In each time step $t$, the agent draws an action $a_t$ as a movement of the current solution $x_t$, that is, $a_t\in \mathcal{A} = \mathbb{R}^n$. Then, $x_t$ is updated by rounding $x_t+a_t$, that is, $x_{t+1} = [x_t + a_t]$ and $x_{t+1}$ is integral but not guaranteed to satisfy the constraints (\ref{mip:cons}). The transition of state is deterministic and $s_{t+1} = (\text{Flat}(A), b, \tilde{x}_{t+1}, x_{t+1} ,I)$ (for IP, $s_{t+1} = (\text{Flat}(A), b, \tilde{x}_{t+1}, x_{t+1} )$), where $\tilde{x}_{t+1}$ is the projection of $x_{t+1}$ to the feasible region defined by (\ref{mip:cons}).

\textbf{Reward.} As our goal is to find a feasible solution, we use the total violation of constraints (\ref{mip:cons}) as the reward, that is, $r_t = -\|Ax_t -b\|$. As in each step, $x_t$ is guaranteed to be integral, it is reasonable to only consider the violation of constraints (\ref{mip:cons}) rather than constraints (\ref{mip:int}).

\subsection{Policy learning}
\label{sec: policy-learning}
In this part, we demonstrate our RL model. To train the policy network, we use actor-critic algorithms with proximal policy optimization (PPO) as proposed in \cite{schulman2017proximal}. Moreover, two different policy network structures: the classic MLP and a novel CNN are adopted for the policy network of SFP-MLP and SFP-CNN, respectively.

\textbf{Actor-Critic with PPO.}
For policy gradient methods, we use PPO for actor-critic. PPO is a family of policy gradient methods that use different surrogate objective functions compared to standard policy gradient methods. In particular, we use the clipped surrogate objective method that proposed in \cite{schulman2017proximal}, that is, we use the following objective function while training the policy network:
\begin{equation}
\centering
\label{eq:ppo}
L^{CLIP}(\theta) = \hat{\mathbb{E}}_t[\min (r_t(\theta)\hat{A}_t, clip(r_t(\theta), 1-\epsilon, 1+\epsilon)\hat{A}_t)],
\end{equation}
where $\hat{A}_t$ is an estimator of the advantage function at time step $t$. We refer to \cite{schulman2017proximal} for more details about the PPO methods. 

\begin{figure}[!ht]
\centering
\includegraphics[width=0.78\linewidth]{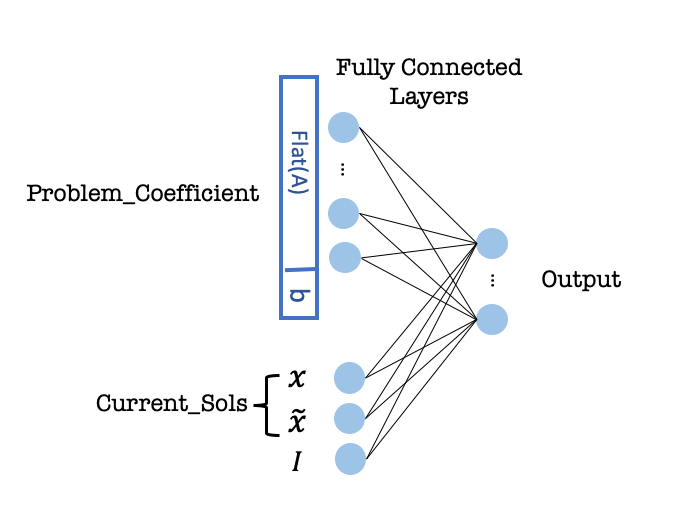}
\caption{Policy Network Structure of SFP-MLP.} 
\label{fig:mlp}
\end{figure}

\begin{figure}[!ht]
\centering
\includegraphics[width=\linewidth]{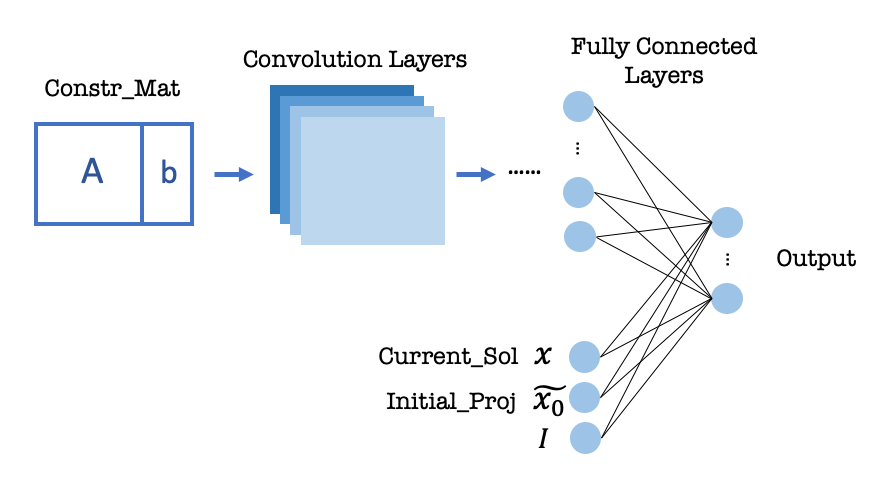}
\caption{Policy Network Structure of SFP-CNN.} 
\label{fig:cnn}
\end{figure}

\textbf{Policy network structure.}
\begin{itemize}
    \item \textbf{SFP-MLP.} As illustrated in Figure~\ref{fig:mlp}, in the SFP-MLP model, the policy network is an MLP that takes the state vector as input. The state vector consists of a flattened constraint matrix $A$, constraint vector $b$, current solution $x_t$, and its projection $\tilde{x}_t$, as well as the integral indicator vector $I$.
    \item \textbf{SFP-CNN.} To better characterize the constraint matrix $(A,b)$ in IP/MIP defined in (\ref{mip}), we use a CNN to capture the policy $\pi_\theta(a|s)$, instead of an MLP. The network structure is illustrated in Figure \ref{fig:cnn}. The inputs of the neural network include two parts, where $\texttt{\textbf{Constr\_Mat}}$ represents the constraint matrix $[A,b]$ and $\texttt{\textbf{Current\_Sols}}$ represents the current solution $x$ and the initial solution $x_0$. In the case of MIPs, there is an additional input $\texttt{\textbf{I}}$ that indicates the index of the integral variables. Unlike using MLP to fit the policy, we keep using the initial solution (which is the rounded optimal solution of a continuous relaxation) instead of the projection of the current solution. Therefore, there is no need to calculate the projection at each time step, which makes our method more computationally efficient compared to using MLP for policy fitting. 
\end{itemize}

\begin{figure*}
\centering
\begin{subfigure}[!ht]{0.3\textwidth}
     \centering
     \includegraphics[width=\textwidth]{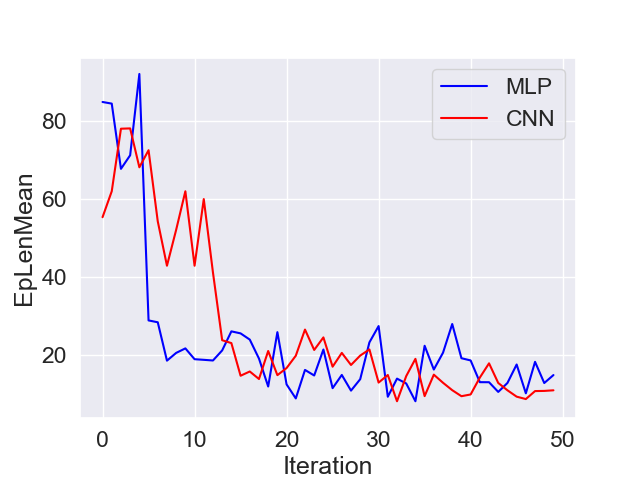}
     \centering
     \includegraphics[width=\textwidth]{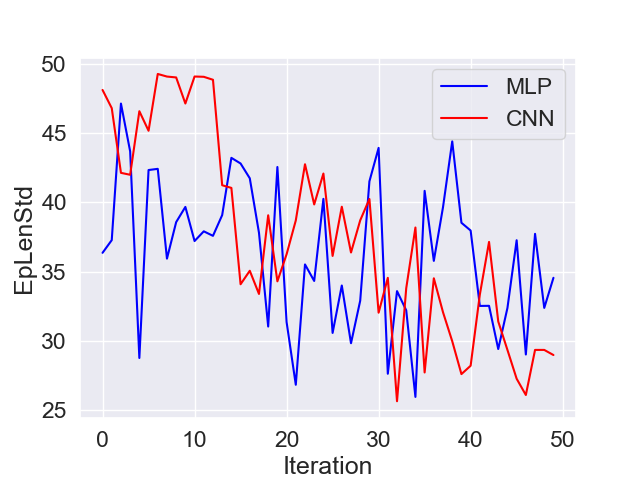}
     \caption{5-dim, 6 constraints}
     \label{fig:5-6}
 \end{subfigure}
 \hfill
 \begin{subfigure}[!ht]{0.3\textwidth}
     \centering
     \includegraphics[width=\textwidth]{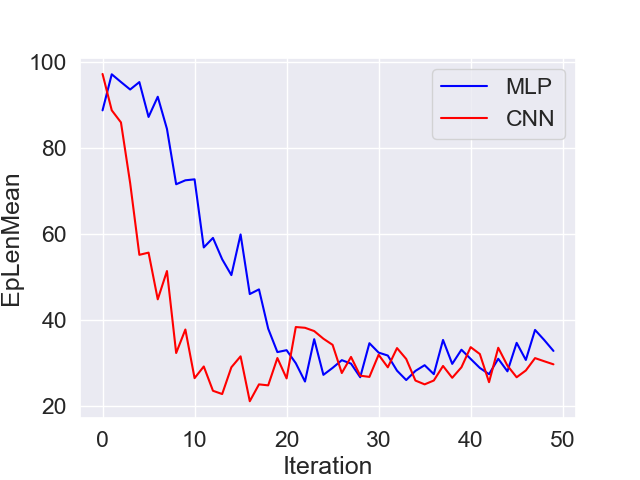}
     \includegraphics[width=\textwidth]{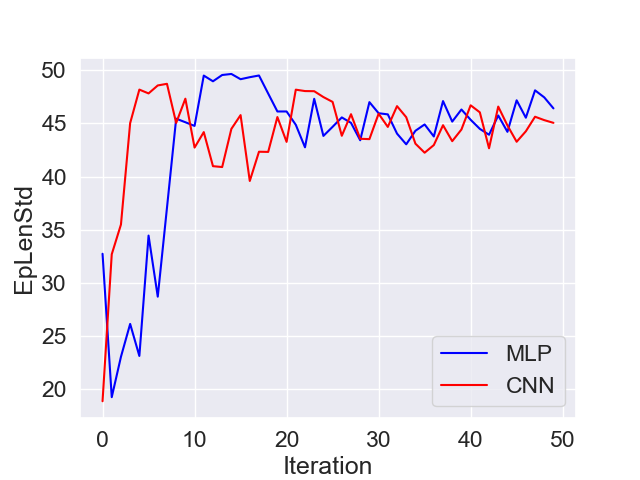}
     \caption{7-dim, 9 constraints}
     \label{fig:7-9}
 \end{subfigure}
 \hfill
 \begin{subfigure}[!ht]{0.3\textwidth}
     \centering
     \includegraphics[width=\textwidth]{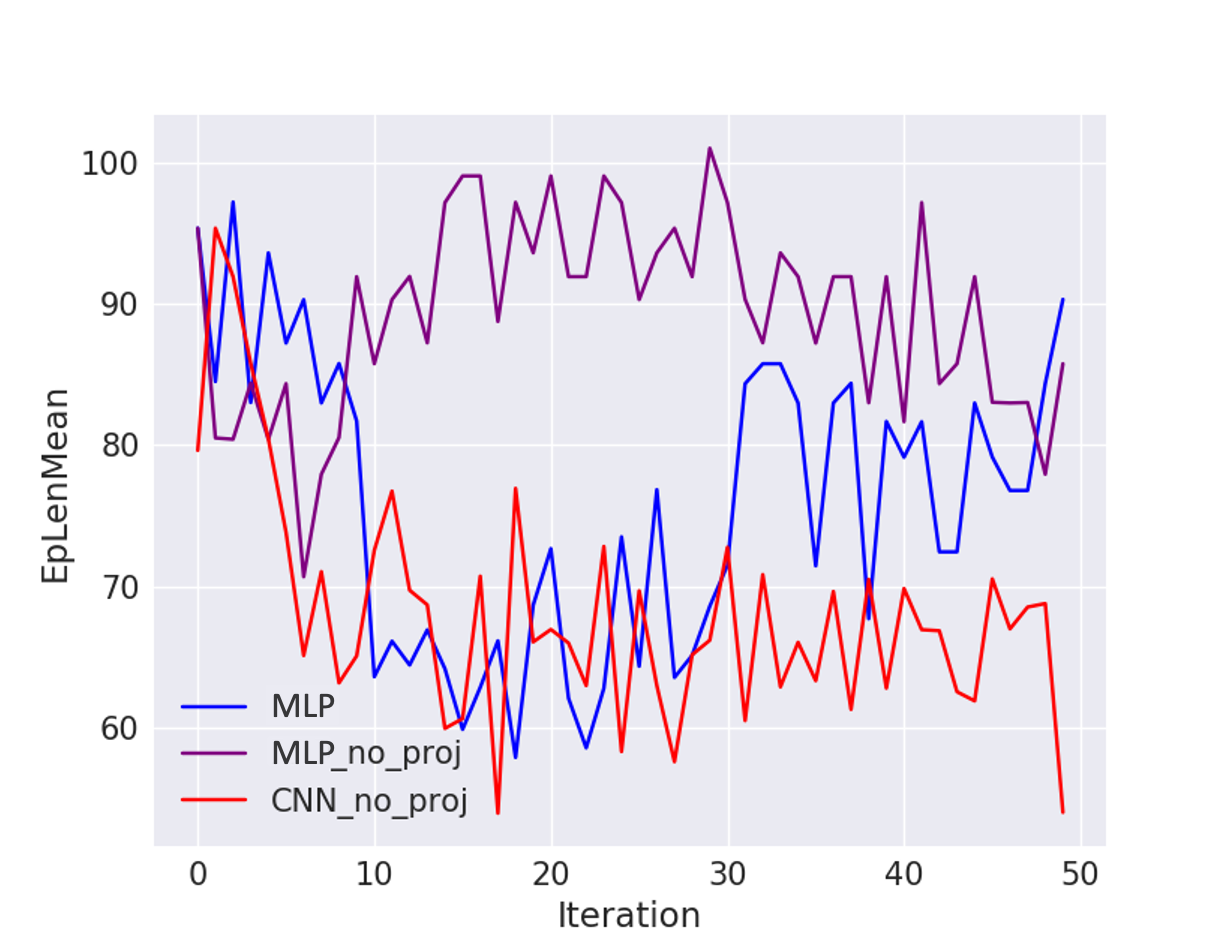}
     \includegraphics[width=\textwidth]{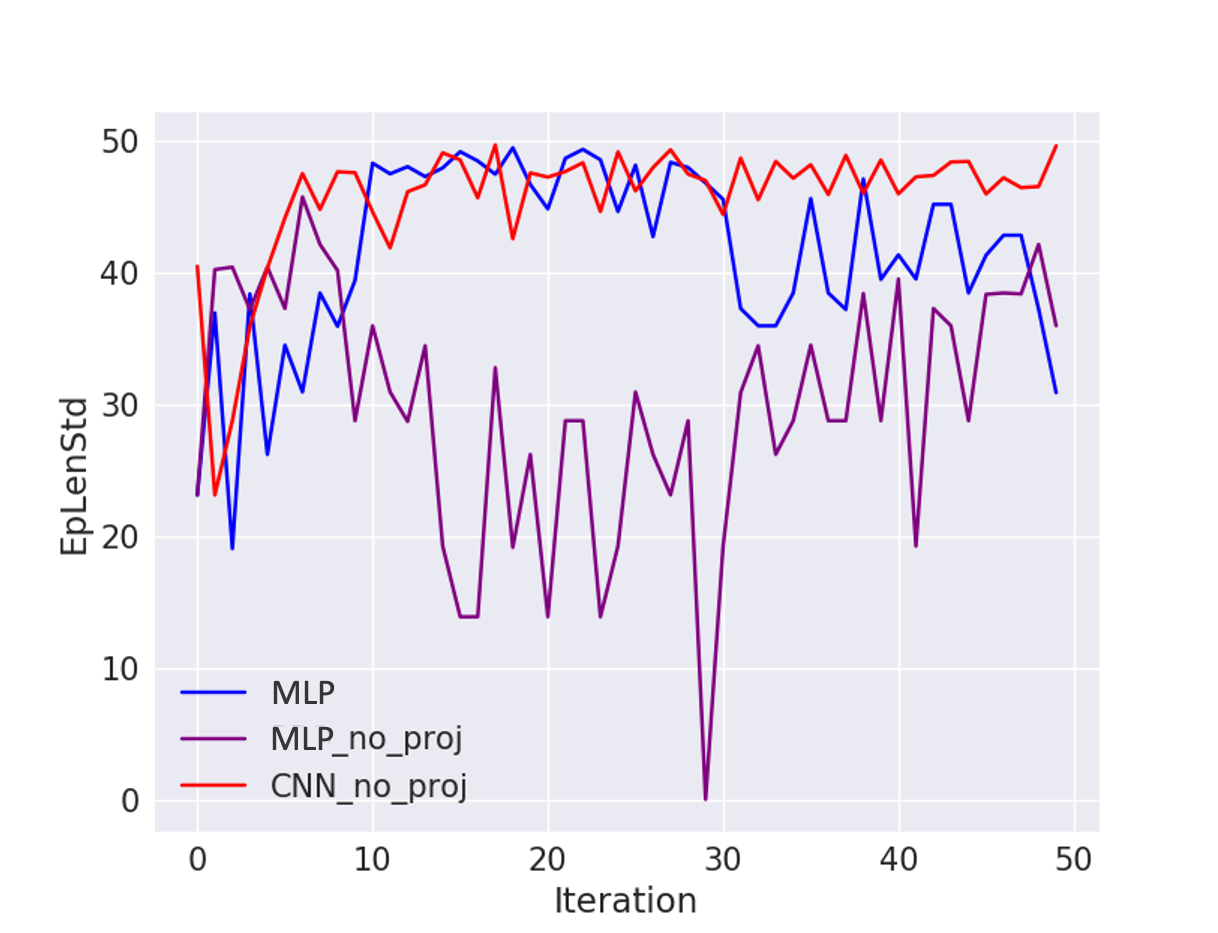}
     \caption{9-dim, 18 constraints}
     \label{fig:9-18}
 \end{subfigure}
 
    \caption{Comparison of SFP-MLP and SFP-CNN. In each of the subfigures (a)--(c), we demonstrate the performance comparison of SFP-MLP and SFP-CNN in the average number of steps (denoted by EpLenMean) and the standard deviation of the number of steps (denoted by EpLenStd) in each iteration. Note that the agent stops if a feasible solution is found or if it reaches the maximum number of steps (100). Therefore, in our setting, it is not always better to have a lower variance in the number of steps. Since the agent interacts with the environment for at most 100 steps, a poorly performed model might have EpLenStd $ = 0$ if it fails to find any feasible solution. An ideal model has both a low EpLenMean and a low EpLenStd. A model is dominated by another model if it has a higher EpLenMean and a lower EpLenStd.}
    \label{fig: mlp vs cnn}
\end{figure*}


\section{Evaluation}
\label{sec:exp}
\subsection{Experiment Design}
\textbf{Simulator Design and Experiment Setting.}
We randomly generate a set of IP and MIP problem instances of form \eqref{mip} based on the parameters listed in Table \ref{tb: problem generator}. The problem instances are small, moderate and large in terms of the dimension of the decision variable $n$ and the number of constraints $m$, including a 5-dimensional decision variable with six constraints; 7-dimensional decision variable with nine constraints; and a 9-dimensional decision variable with 18 constraints. We implemented a simulation environment for the SFP agent. In each training iteration, the agent interacts with one of the generated problems starting at the optimal continuous relaxation. For each problem size, the agent is trained for 50 iterations. The maximum number of training steps is 100. The agent either finds a feasible solution within 100 steps or stops at the 100-th step. Hence, step $< 100$ means that a feasible solution has been found, and step $= 100$ means that the agent has failed for this sample. The same problem instances are tested on the original FP algorithm for comparison. 

\begin{table}[!ht]
\centering
\caption{Problem Parameters}
\begin{tabular}{l|ll}
\hline
Parameter & Distribution\\
 \hline
 $A_{ij}$ & randint[-10,10]  \\
 $b$  & $A\xi + \epsilon$, where $\xi_j\sim$ randint[1,10] and\\& $\epsilon_j\sim$ randint[1,10] for all $j = 1,...,n$ \\
 $I$ & randint[0,1]\\
\hline
\end{tabular}

\label{tb: problem generator}
\end{table}

\textbf{Performance Evaluation Metric.}
The reward per iteration measures that the violation of the constraints is not a direct metric of the model performance in the context of our application since our primary goal is to find a feasible solution within a small number of steps. We consider the number of steps to reach a feasible solution as our performance evaluation metric instead of the reward per iteration. In particular, we evaluated the empirical mean and standard deviation of the number of steps, which we denote as \textbf{EpLenMean} and \textbf{EpLenStd}, respectively. Note that in this setting, it is \textit{not} always better to have a \textit{lower} variance of the number of steps. Because our agent interacts with the environment for at most 100 steps, a poorly performed model may have EpLenMean $= 100$ but EpLenStd $ = 0$ if it fails to find any feasible solution. An ideal model has both a low EpLenMean and a low EpLenStd. A model outperforms model with higher EpLenMean and lower EpLenStd because the latter one is stable with more steps to reach a feasible solution. 

\subsection{Empirical Results}
\begin{table*}[!ht]
\centering
\caption{Comparison with Feasibility Pump (IP). }
\begin{tabular}{c|ccc|ccc|ccc}
\hline
\multicolumn{1}{l|}{} & \multicolumn{3}{c|}{$n = 5, m = 6$} & \multicolumn{3}{c|}{$n = 7, m = 9$} & \multicolumn{3}{c|}{$n = 9, m = 18$} \\ \cline{2-10} 
\multicolumn{1}{l|}{} & FP         & MLP        & CNN       & FP         & MLP        & CNN       & FP         & MLP        & CNN        \\ \hline
EpLenMean            & 43.2       & 15.0       & 11.1      & 65.5       & 32.8       & 28.2      & 90.0       & 90.3       & 54.0       \\ \hline
EpLenStd            & 48.6       & 34.6       & 29.0      & 46.8       & 46.4       & 44.2      & 29.7       & 30.9       & 49.6       \\ \hline
EpLenMax            & 100.0      & 100.0      & 100.0     & 100.0      & 100.0      & 100.0     & 100.0      & 100.0      & 100.0      \\ \hline
90 Quant              & 100.0      & 100.0      & 29.5      & 100.0      & 100.0      & 100.0     & 100.0      & 100.0      & 100.0      \\ \hline
10 Quant              & 1.0        & 1.0        & 1.0       & 1.0        & 1.0        & 1.0       & 61.2       & 51.5       & 1.0        \\ \hline
\end{tabular}
\label{tab:fp_comp}
\end{table*}

\begin{table*}[!ht]
\centering
\caption{Comparison with Feasibility Pump (MIP)}
\begin{tabular}{c|ccc|ccc|ccc}
\hline
\multicolumn{1}{l|}{} & \multicolumn{3}{c|}{$n = 5, m = 6$} & \multicolumn{3}{c|}{$n = 7, m = 9$} & \multicolumn{3}{c|}{$n = 9, m = 18$} \\ \cline{2-10} 
\multicolumn{1}{l|}{} & FP         & MLP        & CNN       & FP         & MLP        & CNN       & FP         & MLP        & CNN        \\ \hline
EpLenMean            & 31.9       & 13.1       & 12.7      & 57.0       & 25.0       & 27.2      & 82.7       & 70.5       & 60.8       \\ \hline
EpLenStd            & 45.5       & 32.5       & 31.3      & 48.2       & 42.6       & 42.6      & 36.8       & 46.0       & 48.3       \\ \hline
EpLenMax            & 100.0      & 100.0      & 100.0     & 100.0      & 100.0      & 100.0     & 100.0      & 100.0      & 100.0      \\ \hline
90 Quant              & 100.0      & 100.0      & 100.0     & 100.0      & 100.0      & 100.0     & 100.0      & 100.0      & 100.0      \\ \hline
10 Quant              & 1.0        & 1.0        & 1.0       & 1.0        & 1.0        & 1.0       & 2.0        & 1.0        & 1.0        \\ \hline
\end{tabular}

\label{tab:fp_comp_mip}
\end{table*}
\textbf{Experiment 1: Comparison with FP} In this section, we compare our proposed SFP agents with the classic FP algorithm (Algorithm \ref{alg:fp}. The results are summarized in Table \ref{tab:fp_comp} for the IP problems and Table \ref{tab:fp_comp_mip} for the MIP problems. \texttt{\textbf{EpLenMax}}, \texttt{\textbf{90 Quant}} and \texttt{\textbf{10 Quant}} denote the maximum, 90 quantile and 10 quantile of the number of steps, respectively. The result shows that the SFP agent finds a feasible solution to IP/MIPs faster than the FP algorithm. In addition, SFP-CNN dominates the FP algorithm and SFP-MLP, especially when the problem size increases. Therefore, the SFP agent is much more computationally efficient than the classic FP algorithm. 

\textbf{Experiment 2: MLP and CNN.} In this section, we present the empirical evaluation results for the SFP-MLP and SFP-CNN methods. Figure \ref{fig: mlp vs cnn} shows the training curves of SFP-MLP and SFP-CNN with different problem scales. The results show that SFP-MLP and SFP-CNN are comparable when the problem size is small (Figure \ref{fig:5-6}). Both EpLenMean and EpLenStd decreases when the training iteration increases, which is how an ideal model would perform. SFP-CNN outperforms SFP-MLP in the sense that it converges faster to a lower EpLenMean with comparable EpLenStd when the problem size becomes larger (Figures \ref{fig:7-9} and \ref{fig:9-18}). Figure \ref{fig:9-18} provides an additional comparison of the SFP-MLP without the projection of the previous point. Interestingly, we observe that the performance of SFP-MLP is largely dampened without the projection information, while the performance of SFP-CNN without projection is better than that of SFP-MLP with projection. Thus, SFP-CNN can be more computationally efficient than SFP-MLP with larger problem scales. This also highlights the representational power of the CNN structure to capture hidden information in the constraint matrices.

In summary, the SFP agents find a feasible solution for IP/MIPs efficiently and outperforms the classic FP algorithm. SFP-CNN dominates the classic FP algorithm and SFP-MLP. In addition, SFP-CNN works without the projection of the current point. As noted in Section \ref{sec: background}, it requires solving an optimization problem at each step to obtain the projection. Therefore, SFP-CNN achieves a significant improvement over the original FP method. This highlights the representational power of our proposed CNN structure to capture hidden information in the constraint matrix.

  

\section{Conclusion}

In this work, we propose an SFP, a reinforcement learning-based model for finding feasible solutions to generic IP/MIP problems. 
Unlike the supervised learning scheme, our method does not require the knowledge of a training set that includes feasible solutions. Our model is constructed based on the spirit of a well-known heuristic, the FP. Numerical experiments on different problem scales show that our proposed SFP agents can efficiently find a feasible solution for general IP/MIPs and improve the performance compared to the original FP algorithm. 


To capture the inherent structure of the constraint matrix, we propose a novel CNN structure for the policy network in addition to the MLP policy network. SFP-CNN outperforms SFP-MLP and is more computationally efficient because it does not require finding the projection of the current point at each step. The CNN structure exhibits great representation power for optimization problems.

We suggest several opportunities for future research in this direction. One interesting topic would be to leverage the CNN structure, which can transfer useful information between different problem settings with different numbers of constraints. The other potential direction would be to take the objective value into account and trying to find a feasible solution with better quality.
\label{sec:con}

\bibliographystyle{named}
\bibliography{ijcai21}

\begin{thebibliography}{}

\bibitem[\protect\citeauthoryear{Achterberg and
  Berthold}{2007}]{achterberg2007improving}
Tobias Achterberg and Timo Berthold.
\newblock Improving the feasibility pump.
\newblock {\em Discrete Optimization}, 4(1):77--86, 2007.

\bibitem[\protect\citeauthoryear{Baena and Castro}{2011}]{baena2011using}
Daniel Baena and Jordi Castro.
\newblock Using the analytic center in the feasibility pump.
\newblock {\em Operations Research Letters}, 39(5):310--317, 2011.

\bibitem[\protect\citeauthoryear{Bello \bgroup \em et al.\egroup
  }{2016}]{bello2016neural}
Irwan Bello, Hieu Pham, Quoc~V Le, Mohammad Norouzi, and Samy Bengio.
\newblock Neural combinatorial optimization with reinforcement learning.
\newblock {\em arXiv preprint arXiv:1611.09940}, 2016.

\bibitem[\protect\citeauthoryear{Bertacco \bgroup \em et al.\egroup
  }{2007}]{bertacco2007feasibility}
Livio Bertacco, Matteo Fischetti, and Andrea Lodi.
\newblock A feasibility pump heuristic for general mixed-integer problems.
\newblock {\em Discrete Optimization}, 4(1):63--76, 2007.

\bibitem[\protect\citeauthoryear{Boland \bgroup \em et al.\egroup
  }{2012}]{boland2012new}
Natashia~L Boland, Andrew~C Eberhard, F~Engineer, and Angelos Tsoukalas.
\newblock A new approach to the feasibility pump in mixed integer programming.
\newblock {\em SIAM Journal on Optimization}, 22(3):831--861, 2012.

\bibitem[\protect\citeauthoryear{Boland \bgroup \em et al.\egroup
  }{2014}]{boland2014boosting}
Natashia~L Boland, Andrew~C Eberhard, Faramroze~G Engineer, Matteo Fischetti,
  Martin~WP Savelsbergh, and Angelos Tsoukalas.
\newblock Boosting the feasibility pump.
\newblock {\em Mathematical Programming Computation}, 6(3):255--279, 2014.

\bibitem[\protect\citeauthoryear{Bonami and
  Gon{\c{c}}alves}{2012}]{bonami2012heuristics}
Pierre Bonami and Jo{\~a}o~PM Gon{\c{c}}alves.
\newblock Heuristics for convex mixed integer nonlinear programs.
\newblock {\em Computational Optimization and Applications}, 51(2):729--747,
  2012.

\bibitem[\protect\citeauthoryear{Bonami \bgroup \em et al.\egroup
  }{2009}]{bonami2009feasibility}
Pierre Bonami, G{\'e}rard Cornu{\'e}jols, Andrea Lodi, and Fran{\c{c}}ois
  Margot.
\newblock A feasibility pump for mixed integer nonlinear programs.
\newblock {\em Mathematical Programming}, 119(2):331--352, 2009.

\bibitem[\protect\citeauthoryear{Danna \bgroup \em et al.\egroup }{2005}]{rins}
Emilie Danna, Edward Rothberg, and Claude Le~Pape.
\newblock Exploring relaxation induced neighborhoods to improve mip solutions.
\newblock {\em Mathematical Programming}, 102(1):71--90, 2005.

\bibitem[\protect\citeauthoryear{Dantzig \bgroup \em et al.\egroup
  }{1954}]{dantzig1954solution}
George Dantzig, Ray Fulkerson, and Selmer Johnson.
\newblock Solution of a large-scale traveling-salesman problem.
\newblock {\em Journal of the operations research society of America},
  2(4):393--410, 1954.

\bibitem[\protect\citeauthoryear{Delarue \bgroup \em et al.\egroup
  }{2020}]{delarue2020reinforcement}
Arthur Delarue, Ross Anderson, and Christian Tjandraatmadja.
\newblock Reinforcement learning with combinatorial actions: An application to
  vehicle routing.
\newblock {\em arXiv preprint arXiv:2010.12001}, 2020.

\bibitem[\protect\citeauthoryear{Fischetti and Jo}{2017}]{fischetti2017deep}
Matteo Fischetti and Jason Jo.
\newblock Deep neural networks as 0-1 mixed integer linear programs: A
  feasibility study.
\newblock {\em arXiv preprint arXiv:1712.06174}, 2017.

\bibitem[\protect\citeauthoryear{Fischetti and Lodi}{2003}]{fischetti2003local}
Matteo Fischetti and Andrea Lodi.
\newblock Local branching.
\newblock {\em Mathematical programming}, 98(1-3):23--47, 2003.

\bibitem[\protect\citeauthoryear{Fischetti and
  Salvagnin}{2009}]{fischetti2009feasibility}
Matteo Fischetti and Domenico Salvagnin.
\newblock Feasibility pump 2.0.
\newblock {\em Mathematical Programming Computation}, 1(2-3):201--222, 2009.

\bibitem[\protect\citeauthoryear{Fischetti \bgroup \em et al.\egroup
  }{2005}]{fischetti2005feasibility}
Matteo Fischetti, Fred Glover, and Andrea Lodi.
\newblock The feasibility pump.
\newblock {\em Mathematical Programming}, 104(1):91--104, 2005.

\bibitem[\protect\citeauthoryear{He \bgroup \em et al.\egroup
  }{2014}]{he2014learning}
He~He, Hal Daume~III, and Jason~M Eisner.
\newblock Learning to search in branch and bound algorithms.
\newblock {\em Advances in neural information processing systems},
  27:3293--3301, 2014.

\bibitem[\protect\citeauthoryear{Khalil \bgroup \em et al.\egroup
  }{2016}]{learning_branch}
Elias~Boutros Khalil, Pierre Le~Bodic, Le~Song, George~L Nemhauser, and
  Bistra~N Dilkina.
\newblock Learning to branch in mixed integer programming.
\newblock In {\em AAAI}, pages 724--731, 2016.

\bibitem[\protect\citeauthoryear{Khalil \bgroup \em et al.\egroup
  }{2017}]{khalil2017learning}
Elias Khalil, Hanjun Dai, Yuyu Zhang, Bistra Dilkina, and Le~Song.
\newblock Learning combinatorial optimization algorithms over graphs.
\newblock In {\em Advances in neural information processing systems}, pages
  6348--6358, 2017.

\bibitem[\protect\citeauthoryear{Land and Doig}{2010}]{land2010automatic}
Ailsa~H Land and Alison~G Doig.
\newblock An automatic method for solving discrete programming problems.
\newblock In {\em 50 Years of Integer Programming 1958-2008}, pages 105--132.
  Springer, 2010.

\bibitem[\protect\citeauthoryear{Li \bgroup \em et al.\egroup
  }{2018}]{li2018combinatorial}
Zhuwen Li, Qifeng Chen, and Vladlen Koltun.
\newblock Combinatorial optimization with graph convolutional networks and
  guided tree search.
\newblock In {\em Advances in Neural Information Processing Systems}, pages
  539--548, 2018.

\bibitem[\protect\citeauthoryear{Malandraki and
  Daskin}{1992}]{malandraki1992time}
Chryssi Malandraki and Mark~S Daskin.
\newblock Time dependent vehicle routing problems: Formulations, properties and
  heuristic algorithms.
\newblock {\em Transportation science}, 26(3):185--200, 1992.

\bibitem[\protect\citeauthoryear{Nair \bgroup \em et al.\egroup
  }{2020}]{nair2020solving}
Vinod Nair, Sergey Bartunov, Felix Gimeno, Ingrid von Glehn, Pawel Lichocki,
  Ivan Lobov, Brendan O'Donoghue, Nicolas Sonnerat, Christian Tjandraatmadja,
  Pengming Wang, et~al.
\newblock Solving mixed integer programs using neural networks.
\newblock {\em arXiv preprint arXiv:2012.13349}, 2020.

\bibitem[\protect\citeauthoryear{Nazari \bgroup \em et al.\egroup
  }{2018}]{learn_vrp}
MohammadReza Nazari, Afshin Oroojlooy, Lawrence Snyder, and Martin Takac.
\newblock Reinforcement learning for solving the vehicle routing problem.
\newblock In {\em Advances in Neural Information Processing Systems}, pages
  9860--9870, 2018.

\bibitem[\protect\citeauthoryear{Padberg}{1973}]{padberg1973facial}
Manfred~W Padberg.
\newblock On the facial structure of set packing polyhedra.
\newblock {\em Mathematical programming}, 5(1):199--215, 1973.

\bibitem[\protect\citeauthoryear{Pochet and
  Wolsey}{2006}]{pochet2006production}
Yves Pochet and Laurence~A Wolsey.
\newblock {\em Production planning by mixed integer programming}.
\newblock Springer Science \& Business Media, 2006.

\bibitem[\protect\citeauthoryear{Sawik}{2011}]{sawik2011scheduling}
Tadeusz Sawik.
\newblock {\em Scheduling in supply chains using mixed integer programming}.
\newblock Wiley Online Library, 2011.

\bibitem[\protect\citeauthoryear{Schulman \bgroup \em et al.\egroup
  }{2017}]{schulman2017proximal}
John Schulman, Filip Wolski, Prafulla Dhariwal, Alec Radford, and Oleg Klimov.
\newblock Proximal policy optimization algorithms.
\newblock {\em arXiv preprint arXiv:1707.06347}, 2017.

\bibitem[\protect\citeauthoryear{Tang \bgroup \em et al.\egroup
  }{2020}]{tang2020reinforcement}
Yunhao Tang, Shipra Agrawal, and Yuri Faenza.
\newblock Reinforcement learning for integer programming: Learning to cut.
\newblock In {\em International Conference on Machine Learning}, pages
  9367--9376. PMLR, 2020.

\end{thebibliography}

\end{document}